\documentclass[10pt, a4paper]{article}

\usepackage[]{lrec-coling2024} 
\usepackage{multirow}
\usepackage{algorithm}
\usepackage{algorithmic}
\usepackage{booktabs}
\usepackage{subcaption}
\usepackage{newfloat}
\usepackage{listings}
\floatstyle{ruled}
\newfloat{listing}{tb}{lst}{}
\floatname{listing}{Listing}

\title{Topic Detection and Tracking with Time-Aware Document Embeddings}

\name{Hang Jiang, Doug Beeferman, Weiquan Mao, Deb Roy} 

\address{Massachusetts Institute of Technology, Stanford University \\
75 Amherst St, Cambridge, MA 02139 \\
         \{hjian42, dougb5, dkroy\}@mit.edu, mwq@stanford.edu\\}

\abstract{
The time at which a message is communicated is a vital piece of metadata
in many real-world natural language processing tasks such as Topic Detection
and Tracking (TDT).   TDT systems aim to cluster a corpus of news articles
by event, and in that context, stories that describe the same event are 
likely to have been written at around the same time.  
Prior work on time modeling for TDT takes this into account, but
does not well capture how time
interacts with the semantic nature of the event.  For example, stories about a
tropical storm
are likely to be written within a short time interval, while stories
about a movie release may appear over weeks or months.  In our work, we
design a neural method that fuses temporal and textual information
into a single representation of news documents for event detection.
We fine-tune these time-aware document embeddings with a triplet loss
architecture, integrate the model into downstream TDT systems, and
evaluate the systems on two benchmark TDT data sets in English. In
the retrospective setting, we apply clustering algorithms to the
time-aware embeddings and show substantial improvements over
baselines on the News2013 data set. In the online streaming setting, we add our document
encoder to an existing state-of-the-art TDT pipeline and demonstrate that
it can benefit the overall performance.  We conduct ablation studies on
the time representation and fusion algorithm strategies, showing that our
proposed model outperforms alternative strategies.  Finally, we probe
the model to examine how it handles recurring events more
effectively than previous TDT systems.
\\ \newline  }

\begin{document}

\maketitleabstract

\section{Introduction}

Following emerging news stories is crucial to making real-time decisions on important political and public safety matters. Amid the COVID-19 pandemic, for instance, media platforms and government agencies need to identify the emergence of misinformation in time and take action to protect the safety of the public.  As humans cannot read all of the articles produced by the news media, automatic clustering of news articles into real-world events is needed to make this work tractable.
Computational tools for this task are useful for organizing 
not just news articles, but also scientific papers, microblogs, online reviews, forum messages, and social media posts.

\citet{allan1998topic} first introduced the topic detection and tracking (TDT) framework to
address these needs.  In this paper, we focus on the event detection part of TDT, where events are defined as real-world news stories and we want to categorize news articles into these events. 
Researchers have proposed both topic modeling and clustering-based methods for this task, in both retrospective and online streaming settings. 
In the retrospective setting, researchers process the news documents altogether, whereas in the online setting they are processed one by one as they appear in the stream. 

Most real-world documents have a timestamp, the time at which it was written, spoken, or shared.
TDT modelers use this metadata, at a minimum, to exploit the temporal locality inherent in news events.
But beyond that, researchers tend to focus on improving text representations \cite{hu2017adaptive,staykovski2019dense,saravanakumar2021event,fan2021clustering} 
instead of time representations.  The time model is typically parameter-free or 
low capacity, and unconnected to the text.  
This overlooks important ways that the time may interact with the topic of the
article to influence where it falls on the calendar and how long it lasts as an event.
For example, stock reports happen
daily, but corporate earnings releases may happen quarterly and be discussed over several days.
Stories about a weather event are likely to be written over a short time period, while stories
a movie release may be discussed over weeks or months. 

In this paper, we address these gaps with a unified neural time-text model that 
fuses temporal and semantic embeddings to represent news documents.   
Our contributions are as follows:

\begin{itemize}
     \item We propose a time-aware neural document embedding method that can be applied to topic detection and tracking and other NLP tasks.
     \item We build two TDT pipelines based on our time-aware model, for retrospective and online event clustering tasks respectively, achieving state-of-the-art performance in both settings for the corpus used as a benchmark by recent prior work.  Importantly, our retrospective model is free of the TF-IDF features that are needed by similar systems, allowing it to be adapted
     to new domains more easily.
     \item We conduct an ablation analysis on our time representation.  We find that sinusoidal positional embedding outperforms two alternatives, learned positional embedding and Date2Vec, to encode timestamps for TDT. 
     \item We analyze the event predictions of our model and baseline methods and find that our model can better handle recurring events that pose challenges to previous TDT systems.
\end{itemize}

\section{Related Work}

Topic detection and tracking (TDT) program \cite{Liu2009,fiscus2002topic} focuses on building algorithms to organize multilingual, news oriented textual materials from the Internet. Apart from news, TDT techniques are also widely used in processing social media data \cite{xiong2022collective}. Traditionally, some researchers \cite{allan1998topic,yang1998study,xu2019research,liu2020mapping} have focused on applying different topic models (e.g., LDA) for TDT. Other researchers \cite{hatzivassiloglou2000investigation,allan2003flexible,dai2010two,li2020topic} explored the use of text clustering algorithms on sparse features and word embeddings for TDT.

Recent approaches to TDT have explored both sparse and dense features. 
\citet{miranda2018multilingual}
proposes an online clustering method that represents documents with TF-IDF features, 
and demonstrates high performance on a benchmark news 
article data set.  Building on this work, \citet{staykovski2019dense} 
compares sparse TF-IDF features with dense Doc2Vec representations, showing a sizeable 
improvement on the standard data set according to the BCubed evaluation metric.
\citet{saravanakumar2021event} is the first to include BERT contextual 
representations for the task and achieves further improvement. Specifically, they fine-tune 
an entity-aware BERT model on an event similarity task with a triplet loss function. 
They generate triplets for each document using the batch-hard regime \cite{hermans2017defense}. 
In each document in a mini-batch, they mark documents with the same label as positive examples and different labels as negative examples. The hardest positive (biggest positive-anchor document distance) and negative (smallest anchor-negative document distance) examples are picked per anchor document to form a triplet. The entity-aware BERT model is trained to make the embedding distance between anchor and positive documents
closer than anchor and negative documents. Overall, this fine-tuning process 
effectively improves the contextual embedding for the overall TDT system. \citet{santos2022simplifying} simplified the multilingual news clustering process with multilingual document embeddings. Recent studies have also leveraged large language models (LLMs) for text clustering \cite{zhang2023clusterllm,viswanathan2023large} and story discovery \cite{yoon2023unsupervised,yoon2023scstory}. This paper focuses on comparing our method against previous embedding-based methods with small language models such as BERT.

TDT systems vary in how they model the timestamps of the news stories.  Some online approaches combine the time element implicitly by sorting documents in chronological order, dividing them with time slicing, and processing each slice \cite{allan1998topic,yang1998study,dai2010two,hu2017adaptive}.  Other work uses decay functions to extract sparse time features \cite{yang1998study,brants2003system,li2005probabilistic,he2010keep,ribeiro2017unsupervised,miranda2018multilingual,saravanakumar2021event}. However, none of the previous work has used temporal embeddings to represent time for the TDT task. This work aims to introduce a popular temporal embedding method from \citet{devlin2018bert} to TDT such that the fused document embedding contains both temporal and semantic information for clustering.  
  
\section{Methodology}

In this section we
propose a novel method called T-E-BERT to encode news documents by fusing text and time information. We adopt a triplet loss function to train the model on the event similarity task and integrate the fine-tuned model into both retrospective and online TDT pipelines. 

\subsection{The Proposed Model}

\begin{figure}[!ht]
    \centering
    \includegraphics[width=0.4\textwidth]{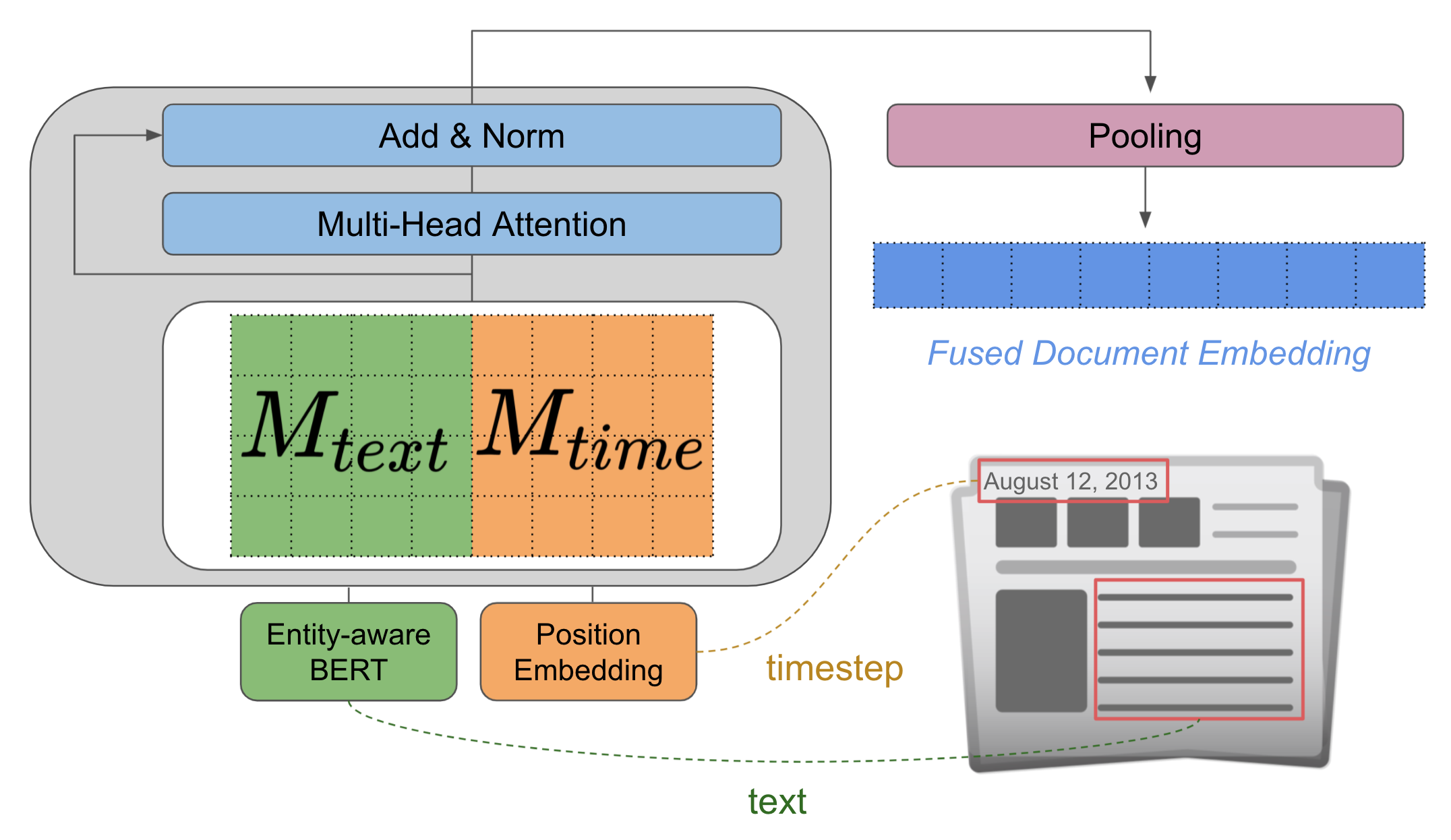}
    \caption{The proposed T-E-BERT model.}
    \label{fig:tebert}
\end{figure}

\subsubsection{T-E-BERT Encoder} We design a simple time-text encoder (``T-E-BERT'') based on entity-aware BERT to combine textual and temporal information to encode news documents (shown in Figure \ref{fig:tebert}). 
        Following \citet{saravanakumar2021event}, we add an entity presence-absence embedding layer to enhance BERT's entity awareness, which can improve the text representation of news events.  We use all the hidden output from the last layer as the textual matrix $M_{text}$\footnote{$M_{text}$\ has a shape of [sequence length, hidden size].}. To represent time, we convert the date time in each document into a time step (e.g. the number of days from the earliest date in the data) and transform the time step into a temporal embedding with the sinusoidal position encoding method introduced by \citet{vaswani2017attention}. The temporal embedding is repeated ``sequence length'' times to generate $M_{time}$, which is the same shape as the $M_{text}$. Afterwards, we introduce a fusion module based on multi-head attention to transform the concatenation of text and time embeddings into a text-time matrix. At last, the fused matrix is fed into the pooling layer to generate a news document embedding. The model is trained on a event similarity task to learn how to combine text and time information before this is used for downstream TDT tasks. 

\subsubsection{Fine-tuning} 
\begin{figure}[!ht]
    \centering
    \includegraphics[width=0.5\textwidth]{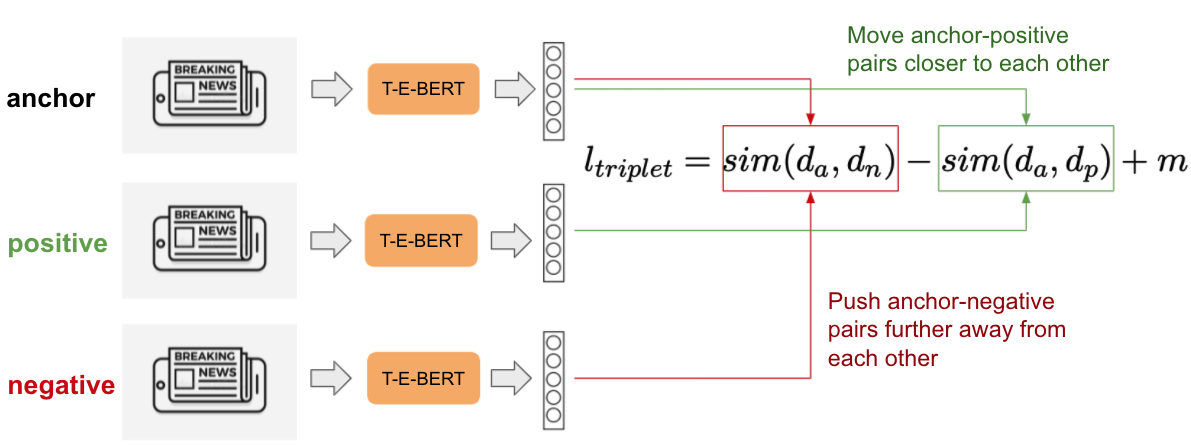}
    \caption{This figures demonstrates the fine-tuning procedure of T-E-BERT on news event triplets.}
    \label{fig:triplet_loss}
\end{figure}

We follow the fine-tuning procedure suggested by \citet{saravanakumar2021event} to adapt the triplet network structure \cite{hoffer2015deep} and fine-tune our T-E-BERT model on the event similarity task. The task aims to tune the model such that it can make the temporal-semantic similarity between same events smaller than the temporal-semantic similarity between different events. In Figure \ref{fig:triplet_loss}, we demonstrate how the training paradigm works. Given an anchor document $d_a$, we sample a positive document $d_p$ (from the same event as $d_a$) as well as a negative document $d_n$ (from a different event). We compute the triplet loss function as follows:

\begin{equation}
    L_{triplet} = sim(d_a, d_n) - sim(d_a, d_p) + m
\end{equation}
where $sim$ is the cosine similarity function and $m$
is the hyper-parameter margin.

\subsection{Retrospective TDT Pipeline}
The retrospective TDT pipeline is simple, consisting of a document encoder and a clustering module. At the document encoding step, we concatenate the title and the body of the news articles to form the input text.
We then replace the original TF-IDF encoder \cite{yang1998study,allan1998topic,schultz1999topic} with the fine-tuned T-E-BERT to vectorize the news documents, considering both their timestamps and texts. These vectors are directly fed into a clustering algorithm to produce event clusters. We choose the HDBSCAN clustering algorithm\footnote{https://hdbscan.readthedocs.io/} \cite{mcinnes2017hdbscan} for two reasons. First, HDBSCAN does not require the number of clusters as a hyperparameter,
which is unknown to a TDT system in a real-world deployment.  Second, HDBSCAN shows strong empirical performance in our experiments, even compared with K-Means and agglomerative clustering algorithms using the true number of clusters. 

\subsection{Online TDT Pipeline}
\begin{figure*}[!ht]
    \centering
    \includegraphics[width=0.8\textwidth]{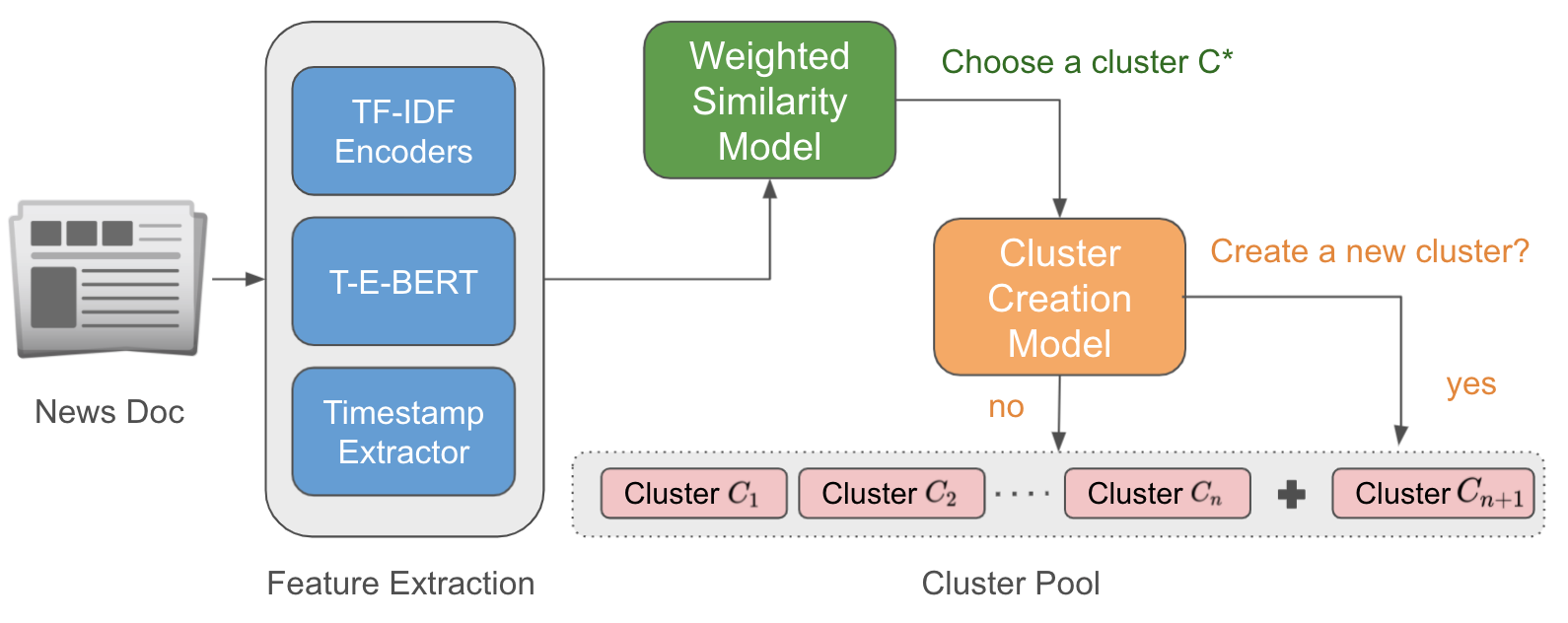}
    \caption{Overview of the online TDT pipeline.}
    \label{fig:tdt_online}
\end{figure*}

We follow the previous work \cite{miranda2018multilingual,saravanakumar2021event} to adopt a variant of the streaming K-means algorithm (Figure \ref{fig:tdt_online}) with a few key changes. This system consists of three main components: (1) a document encoder, (2) a document-cluster weighted similarity model, (3) a cluster creation model. At any point time $t$, let $n$ be the number of clusters in the cluster pool. For any input document, we assume it belongs to a single event cluster. We first represent this document with a set of vectors including 9 TF-IDF sparse vectors, one dense vector from T-E-BERT, and one time sparse vector \cite{saravanakumar2021event}. 
After the document representation is extracted, we
use the document-cluster weighted similarity model to find the best matching event cluster $C^{*}$ from the cluster pool. Finally we use a cluster creation model to decide whether a new cluster is needed. If the cluster $C^{*}$ is predicted to be a good fit for the new document, we add the document to this cluster. Otherwise, we create a new cluster containing this document and add the cluster into the cluster pool.

We make two major changes to previous methods \cite{miranda2018multilingual,saravanakumar2021event}. First, we switch the weighted similarity model from SVM-rank to a linear model with MarginRankingLoss. This change boosts the performance of the weighted similarity model\footnote{We also experimented with SVM-triplet but our implementation performed worse than their reported performance \cite{saravanakumar2021event}.}. Second, we sample balanced positive and negative examples to train the cluster creation model. This alleviates the issue of data imbalance because only 5\% of the training data points contain a positive label (create a new cluster). 

\section{Experiments}

\subsection{Data Sets}
We use two data sets for our experiments, which we refer to as News2013 and TDT-1. 
News2013 refers to the English portion of the multilingual news data set produced in \citet{miranda2018multilingual}, which in turn derives from \citet{rupnik2016news}. The News2013 data set contains a train set from 2013-12-18 to 2014-11-02, and a temporally disjoint test set from 2014-11-02 to 2015-08-25.  TDT-1 refers to the pilot data set for the TDT initiative.  We follow \citet{saravanakumar2021event} to generate its train set from 1994-07-09 to 1995-06-30 and a temporally overlapping test set from 1994-07-04 to 1995-06-28\footnote{We follow their event splits, but get different train and test instances from \citet{saravanakumar2021event}.}. Dataset statistics are shown in Table \ref{tab:datasets}. 

\begin{table}[!ht]
\begin{center}
\setlength{\tabcolsep}{3pt}
\footnotesize
\begin{tabular}{lcccc}
    \toprule
      \multirow{2}{*}{\bf Dataset}&\multicolumn{2}{c}{\bf Train} & \multicolumn{2}{c}{\bf Test}\\
      &$|D|$&$|E|$&$|D|$&$|E|$\\
      \midrule
     News2013{\scriptsize~\cite{miranda2018multilingual}}&12,233&593&8,726&222\\
     TDT-1 {\scriptsize~\cite{allan1998topic}}&899&13&654&12\\
     \bottomrule
\end{tabular}
   \caption{Characteristics of the two TDT datasets. $|D|$ and $|E|$ denote the number of documents and topic events, respectively.}
    \label{tab:datasets}
\end{center}
\end{table}

\begin{table*}[t]
\footnotesize
    \centering
\begin{tabular}{lcccccccc}
    \toprule
    {\bf Model}&\multicolumn{4}{c}{\bf News2013}&\multicolumn{4}{c}{\bf TDT-1}\\
    &Precision&Recall&F1&CN&Precision&Recall&F1&CN\\
    \midrule
    TF-IDF + K-Means&87.00&53.94&66.60&222 &84.23&69.01&75.87&12 \\
    TF-IDF + GAC&71.12&96.06&81.73&222 &42.70&97.35&59.36&12 \\
    BERT + GAC&69.35&88.59&77.79&222 &79.27&89.65&84.14&12 \\
    E-BERT + GAC&71.82&86.82&78.61&222 &76.34&89.84&82.54&12 \\
    SinPE-E-BERT + CM + GAC&83.09&95.04&88.66&222  &80.79&91.96&86.01&12 \\
     \midrule
    TF-IDF + HDBSCAN&88.45&58.58&70.48&301 &93.05&89.34&\textbf{91.16}&11 \\
    BERT + HDBSCAN&81.93&69.22&75.04&208 &83.57&90.95&87.10&12 \\
    E-BERT + HDBSCAN&82.07&70.41&75.79&210 &82.23&90.50&86.17&11 \\
    Date2Vec-E-BERT + CM + HDBSCAN&78.68&88.99&83.52&156 &62.83&58.32&60.49&10 \\
    LearnPE-E-BERT + CM + HDBSCAN&83.33&54.24&65.71&252 &79.51&59.36&67.97&13 \\
    SinPE-E-BERT + CM + HDBSCAN &90.18&89.90&\textbf{90.04}&186 &86.79&90.96&88.83&9 \\
    \bottomrule
    \end{tabular}
   \caption{Retrospective TDT performance comparison of baselines with our T-E-BERT variants. CN stands for the predicted cluster number. We adopt the B-Cubed F measure for precision, recall and F1. Note that CN is automatically determined by HDBSCAN but set to be the gold number of clusters for the K-Means and Group Agglomerative Clustering (GAC) algorithms. GAC and HDBSCAN are deterministic, whereas K-Means is stochastic. We run K-Means five times and pick the best run for the table. The average F1 and standard deviation of the K-Means algorithm on two data sets are $67.86\pm 1.00$ (News2013) and $71.06\pm 4.81$ (TDT-1).}
    \label{tab:retro-tdt-results}
\end{table*}

\subsection{Retrospective TDT Experiments}

We designed three experiments in retrospective TDT setting. First, we compare different representations with clustering algorithms that need the gold number of clusters (Table \ref{tab:retro-tdt-results}). We include two known baseline algorithms based on TF-IDF features including (a) the K-Means algorithm and (b) the augmented Group Average Clustering (GAC) \cite{yang1998study}. We then compare them against different BERT representations on GAC to understand the effect of representations. Second, we conduct experiments to compare these representations (TF-IDF, BERT, E-BERT, three variants of T-E-BERT) with HDBSCAN, which does not require the cluster number as an input. In Table \ref{tab:retro-tdt-results}, we 
also conduct experiments with three time encoding strategies (Date2Vec, LearnPE, SinPE). At last, we run some studies on the time-text fusion strategy and time granularity option. With the most performant time encoding method SinPE, we compare four time-text fusion methods (Table \ref{tab:fusion_strategy}) and five time granularity choices on News2013 (Table \ref{tab:time_granularity}). 

\subsubsection{Experiment Setup} We use the online BatchHardTriplet algorithm to fine-tune BERT models for 1-5 epochs and use the best model. The training adopts a batch size of 32 and a max sequence length 230 to best fit into an 11GB GTX 1080 Ti GPU.

\subsubsection{Time Encoding Strategies}
We explored three ways of encoding temporal information into dense embeddings: (1) Date2Vec; (2) learned position embedding (LearnPE); (3) sinusoidal position embedding (SinPE). 
The first method directly transforms a date-time string into a dense vector. The second and third methods transform a document timestamp into a position by calculating the time span (e.g., in days) between the target document and the earliest document,
generating a position embedding from this relative position. 
Previous studies \cite{vaswani2017attention,wang2020position} show that (2) and (3) perform
similarly for language modeling.   To integrate these time encodings into the TDT system, we use the E-BERT text encoder and apply the concatenation + multi-head-attention fusion method described in the method section (Figure \ref{fig:tebert}). 

\begin{itemize}
    \item \textbf{Date2Vec} is a pre-trained time encoder based on Time2Vec \cite{kazemi2019time2vec} that transforms a date and time into a dense vector, while preserving the time-specific characteristics (progression, periodicity, scale, etc). \citet{kazemi2019time2vec} shows that Date2Vec is able to improve downstream NLP tasks. We use the pre-trained Date2Vec model\footnote{https://github.com/ojus1/Date2Vec} released by \citet{kazemi2019time2vec} in our experiments. We decided to update the Date2Vec module during training because we also tried to freeze the time module, leading to 2\% decrease in F1 on News2013. 
    \item \textbf{Learned position embedding (LearnPE)} is a learned position embedding method used by many Transformer-based models \cite{devlin2018bert,liu2019roberta,radford2019language}. It randomly initializes an embedding layer and updates the look-up table during training. 
    \item \textbf{Sinusoidal position embedding (SinPE)} uses the sine and cosine functions of different frequencies to encode positions \cite{vaswani2017attention}. As shown in the following equations, $i$ is the position index and $j$ is the dimension index. Compared to the learned PE method, this approach assigns a fixed embedding to each position.

\end{itemize}

\begin{equation}
    PE_{(i, 2j)} = sin(i/10000^{2j/d_{model}})
\end{equation}
\begin{equation}
    PE_{(i, 2j+1)} = cos(i/10000^{2j/d_{model}})
\end{equation}

\subsubsection{Time-text Fusion Methods}
We experiment with four methods to fuse the time and content embeddings. We repeat each document's position embedding ``sequence length'' times to form the position matrix $M_{time}$. $M_{text}$ indicates the text matrix from the last layer of the BERT model. $e_f$ indicates the fused document embedding. $\oplus$ indicates addition operation of two matrices. $[M_1, M_2]$ indicates a concatenation operation of two matrices. $\mathrm{POOL}$ indicates the mean pooling layer. $\mathrm{ATT}$ indicates the multi-head attention layer. The third CM equation corresponds to the T-E-BERT diagram in Figure \ref{fig:tebert}. Based on the results with the following four time-text fusion methods on News2013 (Table \ref{tab:fusion_strategy}), the CM fusion method achieves the best results relative to the other three methods (A, AM, and ACM). As a result, \textbf{we decide to adopt the SinPE-E-BERT + CM implementation for T-E-BERT in the later experiments}. 
\begin{itemize}
    \item \textbf{additive (A)}: $$e_{f} = \mathrm{POOL}(M_{text} \oplus M_{time})$$
    \item \textbf{additive + multi-head attention (AM)}: $$e_{f} = \mathrm{POOL}(\mathrm{ATT}(M_{text} \oplus M_{time}))$$
    \item \textbf{concatenate + multi-head attention (CM)}: $$e_{f} = \mathrm{POOL}(\mathrm{ATT}([M_{text}, M_{time}]))$$
    \item \textbf{additive + concatenate + multi-head attention (ACM)}: $$e_{f} = \mathrm{POOL}(\mathrm{ATT}([M_{text}, M_{time}]\oplus M_{time}))$$
\end{itemize}

\begin{table}[ht!]
\scriptsize
    \centering
\begin{tabular}{lcccc}
    \toprule
    {\bf Fusion Method}&Precision&Recall&F1&CN\\
    \midrule
    SinPE-E-BERT + A&88.44&86.54&87.48&190\\
    SinPE-E-BERT + AM&89.03&89.29&89.16&183\\
    SinPE-E-BERT + CM&\textbf{90.18}&\textbf{89.90}&\textbf{90.04}&186\\
    SinPE-E-BERT + ACM&89.50&88.71&89.10&188\\
    \bottomrule
    \end{tabular}
   \caption{This tables shows the effect of four different fusion strategies to retrospective TDT on News2013. HDBSCAN is used in this experiment. CN stands for the predicted cluster number.}
    \label{tab:fusion_strategy}
\end{table}

\subsubsection{Time Granularity}
The unit of time can be a crucial factor in the model's performance. Our work follows the previous work \cite{miranda2018multilingual,saravanakumar2021event} and uses 1 day as the time granularity for
the News2013 data set. We also experimented with hourly, bidaily, weekly (7-day), and monthly (30-day) granularity on this data set 
(Table \ref{tab:time_granularity}). \footnote{Note that in our calculation, we treat the time as an offset from the
earliest document in the data set, and hence a pair of documents that are in the same calendar
month may yield different embedding vectors in a model with monthly granularity; and likewise for weekly.}
Similar experiments led us to choose a 3-month granularity for the TDT-1 data set.

\begin{table}[ht!]
\scriptsize
    \centering
\begin{tabular}{lcccc}
    \toprule
    {\bf Time Granularity}&Precision&Recall&F1&CN\\
    \midrule
    SinPE-E-BERT + hourly&85.95&82.58&84.23&199\\
    SinPE-E-BERT + daily&\textbf{90.18}&\textbf{89.90}&\textbf{90.04}&186\\
    SinPE-E-BERT + bidaily&88.92&88.75&88.84&185\\
    SinPE-E-BERT + weekly&86.55&83.62&85.06&189\\
    SinPE-E-BERT + monthly&82.63&72.45&77.20&200\\
    \bottomrule
    \end{tabular}
   \caption{This table shows the effect of time granularity to retrospective TDT on News2013. SinPE-E-BERT + CM + HDBSCAN is used for this experiment. CN stands for the predicted cluster number.}
    \label{tab:time_granularity}
\end{table}

\subsubsection{Retrospective TDT Results} We make three main observations from Table \ref{tab:retro-tdt-results}. First, 
SinPE-E-BERT + CM + GAC outperforms the other methods based on K-Means and GAC on both data sets. This result provides some evidence that time information is helpful to the retrospective TDT task. Second, SinPE-E-BERT consistently improves E-BERT and BERT on both data sets. With HDBSCAN, it substantially outperforms E-BERT ($+14.25\%$) and BERT ($+15.00\%$) on News2013, which challenges recent TDT systems to detect hundreds of fine-grain topic events over time. It also performs above par compared with E-BERT ($+2.66\%$) and BERT ($+1.73\%$) in TDT-1, where events are more broadly defined and spread in the data. At last, SinPE outperforms the other two ways to encode time information. We find that the training data is insufficient to tune Date2Vec and LearnPE to fit the TDT task, for which a monotonic decrease in event similarity over time is the dominating characteristic for news articles. We will discuss the difference between Date2Vec and SinPE further in the ``Probing Time in T-E-BERT'' section.

It is worth noting that BERT-based methods outperform TF-IDF on News2013, but they slightly underperform TF-IDF on TDT-1. However, our retrospective TDT system with T-E-BERT + HDBSCAN is simple and efficient compared to the TF-IDF counterpart. TF-IDF is more complicated to construct and requires researchers to carefully
choose vocabulary size and stop words for each data set\footnote{We aggregate 9 TF-IDF subvectors \cite{miranda2018multilingual} to create one high dimensional sparse TF-IDF vector for clustering to achieve reported performance.}. In contrast, T-E-BERT has a simple and standard fine-tuning procedure to adapt to new data sets. Moreover, TF-IDF features are high dimensional sparse features, whereas T-E-BERT embeddings are low dimensional dense vectors. That means that it takes significantly longer time for the HDBSCAN algorithm to converge with TF-IDF than T-E-BERT embeddings, especially for large data sets such as News2013. 

\subsection{Online TDT Experiments}

\begin{table*}[t]
\scriptsize
    \centering
\begin{tabular}{lcccccccc}
    \toprule
    {\bf Model}&\multicolumn{4}{c}{\bf News2013}&\multicolumn{4}{c}{\bf TDT-1}\\
    &Precision&Recall&F1&CN&Precision&Recall&F1&CN\\
    \midrule
    \citet{laban2017newslens}&94.37&85.58&89.76&873 &-&-&-&- \\
    \citet{miranda2018multilingual}&94.27&90.25&92.36&326 
    &77.14&90.20&83.16&17 \\
    \citet{staykovski2019dense}& 95.16&93.66&94.41&484 &-&-&-&-\\
    \citet{linger2020batch}&94.19&93.55&93.86&298 &-&-&-&- \\
    \citet{saravanakumar2021event}&94.28&95.25&94.76&276 &-&-&-&- \\
    \midrule
    Ours - TF-IDF + TIME &90.21&95.66&92.86&296 &84.05&93.77&88.65&18 \\
    Ours - TF-IDF + BERT + TIME &93.97&89.46&91.66&359 &84.27&95.27&89.43&18 \\
    Ours - TF-IDF + E-BERT + TIME &93.55&95.35&94.44&315 &84.47&95.82&89.79&18 \\
    Ours - TF-IDF + SinPE-E-BERT + TIME &93.20&97.14&\textbf{95.13}&253 &84.69&97.48&\textbf{90.63}&16 \\
     \bottomrule
    \end{tabular}
   \caption{Online Streaming TDT performance between prior work and our system with different features.}
    \label{tab:stream-tdt-results}
\end{table*}

In online TDT experiments, we integrate different BERT embeddings (BERT, E-BERT, SinPE-E-BERT) into the online TDT system and compare them with our baseline TF-IDF + TIME as well as previous works. The online TDT system has a document encoder, for which
we pick the BERT model with the best performance from the retrospective TDT task (T-E-BERT with SinPE and concatenation + multi-head attention) and train the model with the same configuration for the online TDT system. 

As for weighted similarity and cluster creation models, we follow the steps suggested by \citet{miranda2018multilingual} to create the training data. The only difference is that we balance the training data for the cluster creation model, as suggested by \citet{saravanakumar2021event}. Both the weighting and cluster creation models are trained using 5-fold cross validation to tune hyper-parameters and are applied with the best configuration for inference. The clustering output is evaluated against the gold event cluster labels. To be consistent with
recent previous work \cite{staykovski2019dense,saravanakumar2021event} we use the
B-Cubed measure for evaluation.  B-Cubed is more suitable for cluster evaluation
than the standard F measure, as it favors cluster homogeneity and cluster completeness.

\subsubsection{Online TDT Results} We make three observations from Table \ref{tab:stream-tdt-results}. First, we demonstrate that adding time-text fused embeddings can improve online TDT performance.  TFIDF + SinPE-E-BERT + TIME outperforms TF-IDF + E-BERT + TIME on both data sets. It not only shows better performance in B-Cubed F1 but also alleviates the issue of cluster fragmentation by decreasing the number of generated clusters. Second, our proposed online TDT system shows competitive or better performance compared to previous work with the same set of features on the News2013 dataset. For instance, our TF-IDF + TIME system slightly outperforms \citet{miranda2018multilingual} by 0.5\%. Our TF-IDF + E-BERT + TIME slightly underperforms \citet{saravanakumar2021event} with TF-IDF+E-BERT+TIME by 0.3\%. Finally, we confirm the findings from \citet{saravanakumar2021event} that entity-level information is important to the online TDT task. Entity-aware BERT used in TF-IDF + E-BERT + TIME leverages external entity knowledge to beat TF-IDF + TIME and TFIDF + BERT + TIME on both data sets.

\section{Analysis}

\begin{table*}[!ht]
\tiny
    \centering
\begin{tabular}{ccccccc}
\toprule

\multicolumn{3}{c}{}

& \multicolumn{4}{c}{\begin{tabular}[c]{@{}c@{}}
\textbf{Mean pairwise cosine similarity between stories}  \\ \textbf{(E-BERT → T-E-BERT)}\end{tabular}}                                                                                                                                                                                                    \\ 
\textbf{Cluster (\#id)}                                                                     
& \textbf{Event time span}                                      
& \textbf{Stories in cluster}                                                                                                                                                     & \begin{tabular}[c]{@{}c@{}}Daily stock report\\ (\#663)\end{tabular} & \begin{tabular}[c]{@{}c@{}}Daily stock report\\ (\#734)\end{tabular} & \begin{tabular}[c]{@{}c@{}}Beyoncé video review \\ (\#569)\end{tabular} & \multicolumn{1}{c}{\begin{tabular}[c]{@{}c@{}}Typhoon Neoguri \\ hits Japan \\ (\#1394)\end{tabular}} \\ 
\hline

\begin{tabular}[c]{@{}c@{}}Daily stock report \\ (\#663)\end{tabular}                 & \begin{tabular}[c]{@{}c@{}}Apr 24 (1 day)\end{tabular} & 5                & 0.71 → 0.79                                                           & \multicolumn{1}{l}{}                                                                   & \multicolumn{1}{l}{}                                                   &                                                                                      \\ 
\begin{tabular}[c]{@{}c@{}}Daily stock report\\ (\#734)\end{tabular} & \begin{tabular}[c]{@{}c@{}}Jul 28  (1 day)\end{tabular}   & 13                 & 0.45 → 0.16                                                          & 0.52 → 0.65                                                                              & \multicolumn{1}{l}{}                                                   &                                                                                      \\ 
\begin{tabular}[c]{@{}c@{}}Beyoncé video review \\ (\#569)\end{tabular}               & \begin{tabular}[c]{@{}c@{}}May 18-27 (17 days)\end{tabular} & 16                 & -0.01 → -0.07                                                          & 0.01 → 0.01                                                                            & 0.54 → 0.71                                                             &                                                                                      \\
\begin{tabular}[c]{@{}c@{}}Typhoon Neoguri hits Japan \\ (\#1394)\end{tabular}                       & \begin{tabular}[c]{@{}c@{}}Jul 8-11 (4 days)\end{tabular}  & 33                & -0.07 → -0.05                                                          & -0.05 → -0.03                                                                            & -0.05 → 0.02                                                           & \multicolumn{1}{c}{0.78 → 0.90}                                                     \\ \bottomrule
\end{tabular}
\caption{Selected clusters from the News2013 training set.
The cells with → show the change in 
mean cosine similarity, averaged over all document pairs
from the respective clusters, between the
E-BERT embeddings (left) and the time-infused T-E-BERT embeddings (right).
All dates are in 2014. }
\label{tab:cluster_examples}

\end{table*}

\subsection{Probing Time in T-E-BERT}

\begin{figure}[!ht]
    \centering
    \includegraphics[width=0.45\textwidth]{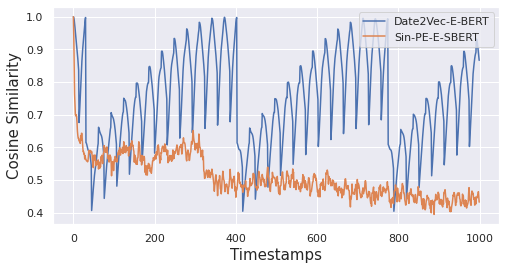}
    \caption{Cosine similarity between the same document with an anchor date and a later date up to 1000 days for our SinPE and Date2Vec models.}
    \label{fig:probe_1000}
\end{figure}

\begin{figure}[htb]
    \centering 
\begin{subfigure}{0.225\textwidth}
  \includegraphics[width=\linewidth]{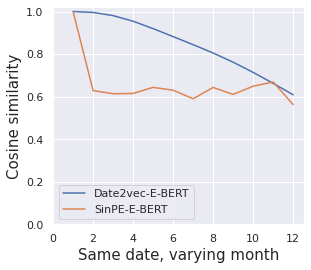}
  \label{fig:1}
\end{subfigure}\hfil 
\begin{subfigure}{0.225\textwidth}
  \includegraphics[width=\linewidth]{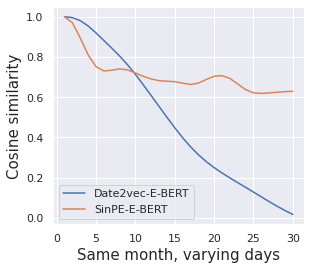}
  \label{fig:2}
\end{subfigure}\hfil 
\caption{Left shows how the cosine similarity changes between the same document with the same date and varying months. Right shows how cosine similarity changes between the same document with the same month and varying days.}
\label{fig:probe_month}
\end{figure}

To understand the effect of time on the overall document embedding, we conduct a simple probing analysis on the Date2Vec-PE-E-BERT and SinPE-E-BERT models. We randomly pick one document from the News2013 dataset and tweak its timestamp from an anchor date to 1000 days later. We compute the cosine similarity $sim(d_0, d_t)$ of the same document between its document embeddings with the anchor date and another date t to generate the Figure \ref{fig:probe_1000}. On one hand, we observe that the similarity monotonously decrease for SinPE-E-BERT, despite some small oscillations in the later days. We suppose that the fluctuations in similarities are due to the limited data used for fine-tuning. News articles are not evenly distributed across time and some dates are associated with more articles. Therefore, it is possible that news articles from some dates are sampled less frequently than others when we dynamically sample triplets for fine-tuning the SinPE-E-BERT model. On the other hand, we notice that the similarity score changes periodically for Date2Vec-PE-E-BERT. Specifically, the similarity for Date2Vec-PE-E-BERT is in a decreasing trend within a month and a week \ref{fig:probe_month}, but the score will be similar for the same dates across months and years \ref{fig:probe_1000}. For instance, Date2Vec-PE-E-BERT thinks a document that the label 2013-12-25 and 2014-12-25 to have a high similarity. However, such a periodic trait is not always beneficial for the TDT task, whereas the span of time from the beginning date is probably the most helpful signal. Therefore, the SinPE-E-BERT model based on position embedding is more suitable for the TDT task than Date2Vec-E-BERT. 

\subsection{Qualitative Analysis}

Both E-BERT and T-E-BERT are fine-tuned such that the embedding vectors produced for
articles within the same training event are moved closer together, but T-E-BERT alone is
sensitive to the timestamp.
Qualitatively, we find that the articles that are impacted the most by T-E-BERT are those 
that belong to
recurring real-world events such as daily stock summaries and financial news updates, 
of which there are many semantically similar clusters in the training corpus.  
Table \ref{tab:cluster_examples} shows four examples of clusters in the News2013
data set; the first two of these are the clusters most impacted by T-E-BERT, both
representing daily stock updates.   We see that the mean pairwise
cosine similarity between the 
articles in the two clusters is much greater for E-BERT (0.45) than for T-E-BERT (0.16);
whereas the within-cluster similarity is in both cases slightly greater for T-E-BERT.

\subsection{Evaluation Metrics}
The B-Cubed metric gives more weight to larger clusters and can obscure the impact
that an algorithm has on smaller clusters. 
In order to weight every cluster equally, we use CEAF-e \cite{luo2005coreference} metric to show that our model performs better on small clusters as well. In addition, other evaluation metrics (Table {\ref{tab:metrics}}) show that our model's superior performance is metric-agnostic.
\begin{table}[ht!]
\scriptsize
    \centering
\begin{tabular}{lccc}
    \toprule
    Metric&Our Best&\citet{saravanakumar2021event}\\
    \midrule
    B-Cubed&\textbf{95.13}&94.76\\
    CEAF-e&\textbf{79.88}&76.93\\
    MUC&99.29&\textbf{99.30}\\
    V Measurey&\textbf{98.04}&97.98\\
    Adjusted Rand Score&\textbf{97.71}&96.26\\
    Adjusted Mutual Info&\textbf{98.20}&97.99\\
    Fowlkes Mallows Score&\textbf{97.77}&96.38\\
    \bottomrule
    \end{tabular}
   \caption{Performance across different evaluation metrics. Our Best represents the best model we have: SinPE-E-BERT + TF-IDF + TIME.  Our model achieves an improvement of 2.95 points on the CEAF-e metric, compared to the best E-BERT + TF-IDF + TIME model in \citet{saravanakumar2021event}.}
    \label{tab:metrics}
\end{table}

\subsection{Truncated Document Length}

We also run ablation studies on triplet mining methods and truncated document length. We compare BatchHardTripletLoss against three other online methods including BatchHardSoftMarginTripletLoss, BatchSemiHardTripletLoss, and BatchAllTripletLoss and four offline methods including Easiest Positive and Easiest Negative(EPEN), Easiest Positive and Hardest Negative (EPHN), Hardest Positive, and Easiest Negative (HPEN), and Hardest Positive and Hardest Negative (HPHN), showing it outperforms the other triplet mining approaches for the retrospective TDT task on News2013. Besides, we find that keeping more than the first few hundred words is of limited help in improving clustering accuracy, confirming the inverted pyramid structure of news articles \cite{po2003news}: the underlying event is usually summarized in the title and the first paragraph, and later paragraphs provide auxiliary information. This suggests that further increasing the sequence length is unlikely to improve the performance substantially. 

\section{Conclusion}
We propose a simple yet effective neural approach to fuse time and text information to create document representations for the TDT task.  We explore different time representations, fusion modules, and time granularities. Our T-E-BERT model SinPE-E-BERT uses sinusoidal positional embeddings to represent timestamps, and entity-aware BERT to represent content. We fine-tune this model with online BatchHardTripletLoss and daily time granularity to achieve state-of-the-art performance on the News2013 benchmark data set. After incorporating our T-E-BERT embeddings in TDT systems, we show superior performance compared to BERT and E-BERT features in both retrospective and online streaming TDT event detection tasks on two benchmark datasets. Finally, we probe our model to show the effectiveness of time module. We also find that SinPE is able to move the document embedding in a desirable direction to better handle recurring events (e.g. stock reports, climate events).

\newpage

\nocite{*}
\section{Bibliographical References}\label{sec:reference}

\bibliographystyle{lrec-coling2024-natbib}
\bibliography{lrec-coling2024-example}



\end{document}